\documentclass[10pt,twocolumn,letterpaper]{article}

\usepackage{wacv}
\usepackage{epsfig}
\usepackage{graphicx}
\usepackage{amsmath}
\usepackage{url}
\usepackage{times}
\usepackage{epsfig}
\usepackage{epstopdf}
\usepackage{amssymb, multirow}
\usepackage{tabularx}
\usepackage{array, booktabs}
\usepackage{subcaption}
\usepackage{threeparttable}
\newcolumntype{P}[1]{>{\centering\arraybackslash}p{#1}}

\def\hlinewd#1{%
\noalign{\ifnum0=`}\fi\hrule \@height #1 %
\futurelet\reserved@a\@xhline} 

\wacvfinalcopy 

\def\wacvPaperID{****} 

\ifwacvfinal\fi
\begin{document}

\title{Vision-based Real Estate Price Estimation}

\author{
Omid Poursaeed,\textsuperscript{1,3} 
Tom{\'a}{\v s} Matera\textsuperscript{3} and 
Serge Belongie\textsuperscript{2,3}\\
\\
\textsuperscript{1} School of Electrical and Computer Engineering, Cornell University\\
\textsuperscript{2} Department of Computer Science, 
Cornell University \quad
\textsuperscript{3} Cornell Tech\\
{\tt\small \{op63,sjb344\}@cornell.edu, tomas@matera.cz}}

\maketitle
\ifwacvfinal\thispagestyle{empty}\fi

\begin{abstract}
Since the advent of online real estate database companies like Zillow, Trulia and Redfin, the problem of automatic estimation of market values for houses has received considerable attention. Several real estate websites provide such estimates using a proprietary formula. Although these estimates are often close to the actual sale prices, in some cases they are highly inaccurate.
One of the key factors that affects the value of a house is its interior and exterior appearance, which is not considered in calculating automatic value estimates. In this paper, we evaluate the impact of visual characteristics of a house on its market value. Using deep convolutional neural networks on a large dataset of photos of home interiors and exteriors, we develop a method for estimating the luxury level of real estate photos. We also develop a novel framework for automated value assessment using the above photos in addition to home characteristics including size, offered price and number of bedrooms. Finally, by applying our proposed method for price estimation to a new dataset of real estate photos and metadata, we show that it outperforms Zillow's estimates.
\end{abstract}

\section{Introduction}
\label{sec1}
The real estate industry has become increasingly digital over the past decade. More than 90\% of home buyers search online in the process of seeking a property \footnote{The Digital House Hunt: Consumer and Market Trends in Real Estate (\url{https://www.nar.realtor/sites/default/files/Study-Digital-House-Hunt-2013-01_1.pdf})}.
Homeowners list their properties on online databases like Zillow, Trulia and Redfin. They provide information on characteristics such as location, size, age, number of bedrooms, number of bathrooms as well as interior and exterior photos. Home buyers, owners, real estate agents and appraisers all need a method to determine the market value of houses. 

A core component of real estate websites like Zillow and Redfin is an automated valuation method (AVM) which estimates the value of a house based on the user-submitted information and publicly available data. The \emph{Zestimate} home valuation is Zillow's estimated market value for houses. It is calculated, using a proprietary formula, for about 110 million homes in the United States. It takes into account factors like physical attributes, tax assessments and prior transactions. The Zestimate has a median error rate of 7.9\%\footnote{This value refers to the reported error rate at the time we started collecting data (June 2016). While the latest reported median error rate of Zestimate is 5.6\% (\url{https://www.zillow.com/zestimate/#acc}), the same approach as what we describe in the paper can be used to decrease the error rate.}, which means half of the Zestimates are closer than the error percentage and half are farther off. Redfin has also released an estimator tool recently that purportedly outperforms Zestimate. It uses massive amounts of data from multiple listing services. Redfin's estimate considers more than 500 data points representing the market, the neighborhood and the home itself to arrive at an estimate for 40 million homes across the United States. It is claimed to have 1.82\% median error rate for homes that are listed for sale, and 6.16\% for off-market homes\footnote{About the Redfin Estimate: \url{www.redfin.com/redfin-estimate} \label{Redfin}}. 

Neither Redfin nor Zillow consider the impact of interior and exterior appearance in their automated valuation methods. However, the visual aspects of a house are key elements in its market value. Home staging companies use this fact to make a home more appealing to buyers.
In view of the importance of design and appearance on the value of a house, in this paper we propose a novel framework for incorporating the impact of interior and exterior design in real estate price estimation. By applying our network to a dataset of houses from Zillow, we evaluate its performance. 

Our contributions can be summarized as follows: 
\begin{itemize}
\item We present the first method which considers the impact of appearance on real estate price estimation.
\item  We elicit luxury-related information from real estate imagery using deep neural networks and crowdsourced data.
\item We release a new dataset of photos and metadata for 9k houses obtained from Zillow. By applying our valuation method to this dataset, we show that it outperforms Zillow's estimates.
\item We release a new, large-scale dataset of 140k interior design photos from Houzz. The images are classified based on their room types: bathroom, bedroom, kitchen, living room, dining room, interior (miscellaneous) and exterior. 
\item We present a qualitative visualization of real estate photos in which images at similar luxury levels are clustered near one another.
\end{itemize}

\section{Related Work}
We now provide an overview of related work, with a focus on automated real estate valuation methods and visual design. We also give a brief overview of machine learning methods and datasets relevant to our approach.

\subsection{Automated Valuation Methods}

Real estate price estimation plays a significant role in several businesses. Home valuation is required for purchase and sale, transfer, tax assessment, inheritance or estate settlement, investment and financing. The goal of automated valuation methods is to automatically estimate the market value of a house based on its available information. Based on the definition of the International Valuation Standards Committee (IVSC), market value is a representation of value in exchange, or the amount a property would bring if offered for sale in the open market at the date of valuation. A survey of real estate price estimation methods is given in \cite{pagourtzi2003real}. In this section, we give an overview of these methods. To our knowledge, none of these methods consider the impact of visual features
on value estimation. 

One of the traditional methods for market valuation is the ``comparables'' model, a form of $k$ nearest neighbors regression. In this model, it is assumed that the value of the property being appraised is closely related to the selling prices of similar properties within the same area. The appraiser must adjust the selling price of each comparable to account for differences between the subject and the comparable. The market value of the subject is inferred from the adjusted sales prices of the comparables. This approach heavily depends on the accuracy and availability of sale transaction data \cite{pagourtzi2003real}.

The problem of price estimation can be viewed as a regression problem in which the dependent variable is the market value of a house and independent variables are home characteristics like size, age, number of bedrooms, etc. Given the market value and characteristics for a large number of houses, the goal is to obtain a function that relates the metadata of a house to its value. There are many bodies of work that apply regression methods to the problem of real estate price estimation. Linear regression models assume that the market value is a weighted sum of home characteristics. They are not robust to outliers and cannot address non-linearity within the data. Another model that is used for price estimation is the hedonic pricing model. It supposes that the relationship between the price and independent variables is a nonlinear logarithmic relation. The interested reader is referred to \cite{pagourtzi2003real}, \cite{benjamin2004mass} and \cite{malpezzi2003hedonic} for an overview of regression analysis for price estimation. Other approaches for price estimation include Artificial Neural Networks \cite{mcgreal1998neural} and fuzzy logic \cite{bagnoli1998theory}. 

Zillow and Redfin use their own algorithms for real estate price estimation. Home characteristics, such as square footage, location or the number of bathrooms are given different weights according to their influence on home sale prices in each specific location over a specific period of time, resulting in a set of valuation rules, or models that are applied to generate each home's Zestimate\footnote{What is a Zestimate? Zillow's Home Value Forecast (\url{http://www.zillow.com/zestimate/})}. Redfin, having direct access to Multiple Listing Services (MLSs), the databases that real estate agents use to list properties, provides a more reliable estimation tool\footnote{About the Redfin Estimate: \url{www.redfin.com/redfin-estimate}}. While Zillow and Redfin do not disclose how they compute their estimates, their algorithms are prone to error, and do not consider the impact of property photos
on the market value of residential properties.
\subsection{Convolutional Neural Networks (ConvNets)}

Convolutional neural networks (ConvNets) have achieved state-of-the-art performance on tasks such as image recognition \cite{krizhevsky2012imagenet,he2016deep,huang2016densely,srivastava2015highway}, segmentation \cite{long2015fully,chen2016deeplab,yu2015multi,zhao2017pyramid}, object detection \cite{girshick2014rich,girshick2015fast,redmon2016you} and generative modeling \cite{goodfellow2014generative,radford2015unsupervised,huang2017stacked,karras2017progressive,poursaeed2017generative} in the last few years. The recent surge of interest in ConvNets recently has resulted in new approaches and architectures appearing on arXiv on a weekly basis. The interested reader is referred to \cite{lecun2015deep} for a review of ConvNets and deep learning.
   
\subsection{Scene Understanding}
One of the hallmark tasks of computer vision is Scene Understanding. In scene recognition the goal is to determine the overall scene category by understanding its global properties. 
The first benchmark for scene classification was the Scene15 database \cite{lazebnik2006beyond}, which contains only 15 scene categories with a few hundred images per class.  The Places dataset is presented in \cite{zhou2014learning}, a scene-centric database containing more than 7 million images from 476 place categories. \cite{yu2015lsun} constructs a new image dataset, named LSUN, which contains around one million labeled images for 10 scene categories and 20 object categories. Table \ref{table:room} shows relevant categories of LSUN, Places and Houzz datasets as well as the number of images in each category. Several categories in the Places dataset are subsumed under the term ``Exterior'': ``apartment building (outdoor)'', ``building facade'', ``chalet'', ``doorway (outdoor)'', ``house'', ``mansion'', ``manufactured home'', ``palace'', ``residential neighborhood'' and ``schoolhouse''.  Miscellaneous indoor classes such as ``stairway'' and ``entrance hall'' are categorized as ``Interior (misc.)''\footnote{While the Houzz dataset contains millions of images in each category, we download and use 20k images in each category.}. 
\begin{table}[!hbp]
\caption{Number of images per room category in different datasets}
\centering
\begin{tabular}{|m{2.8cm} | m{1.47cm} m{0.95cm} m{1.45cm} |}
\hline
   & LSUN & Places & Houzz \\ 
\hline
Living Room & 1,315,802 & 28,842 & 971,512 \\ 
Bedroom & 3,033,042 & 71,033 & 619,180 \\
Dining Room & 657,571 & 27,669 & 435,160 \\
Kitchen/Kitchenette & 2,212,277 & 84,054 & 1,891,946\\
Bathroom & $-$ & 27,990 & 1,173,365 \\
Exterior & $-$ & 25,869 & 868,383 \\
Interior (misc.) & $-$ &  20,000 & 368,293 \\
\hline
\end{tabular}
\label{table:room}
\end{table}
\begin{figure}[!t]
\begin{center}
   \includegraphics[width=\linewidth]{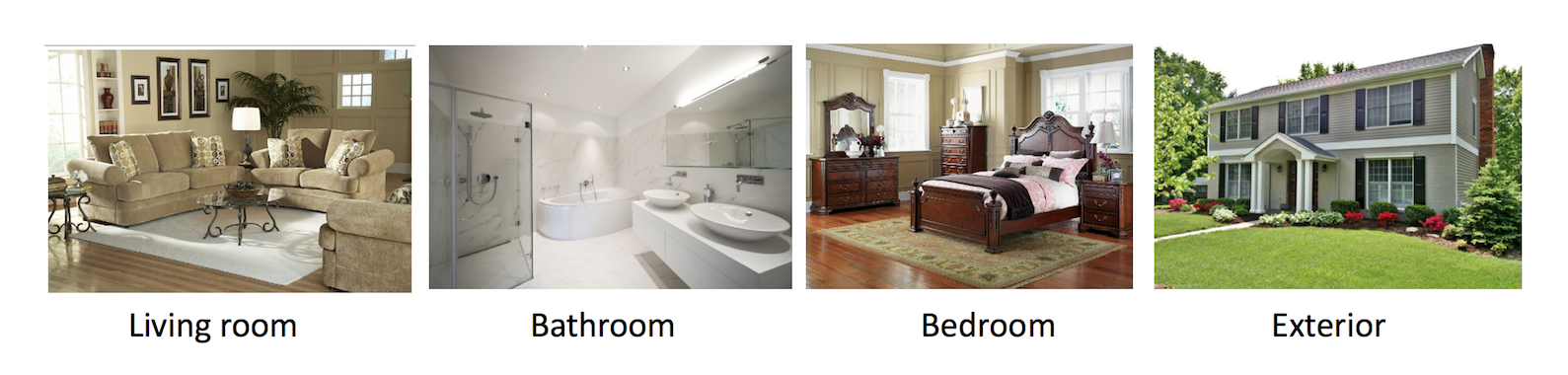}
   \includegraphics[width=\linewidth]{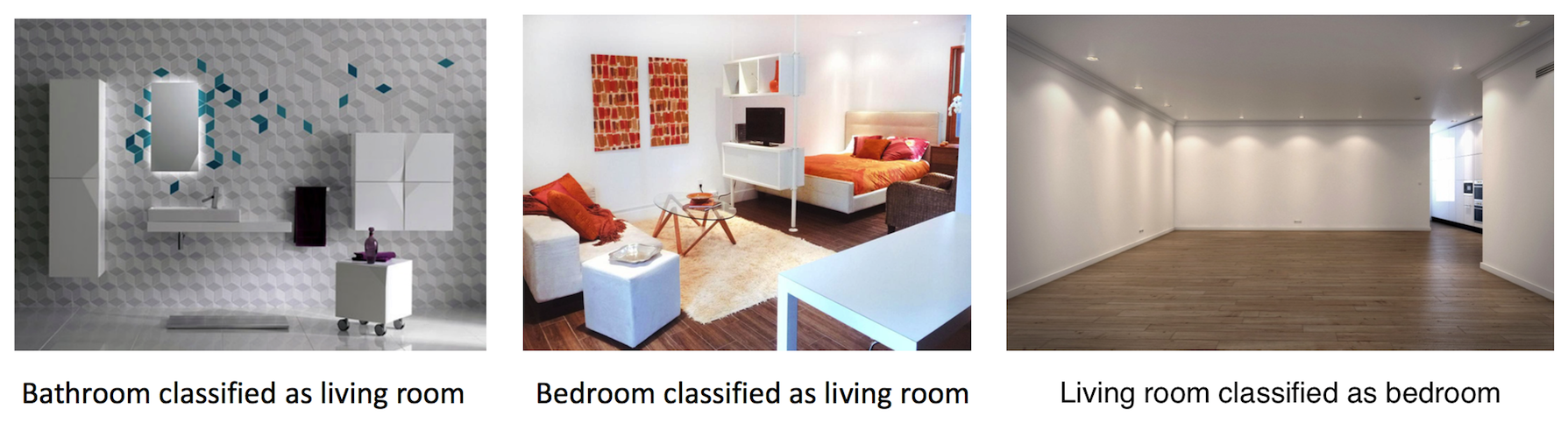}
\end{center}
   \caption{Examples of correctly and incorrectly classified pictures. The first row illustrates images classified correctly, and the second row represents wrongly classified photos. } 
\label{fig:room_classifier}
\end{figure}
\begin{figure*}[!t]
\begin{center}
   \includegraphics[width=0.85\linewidth] {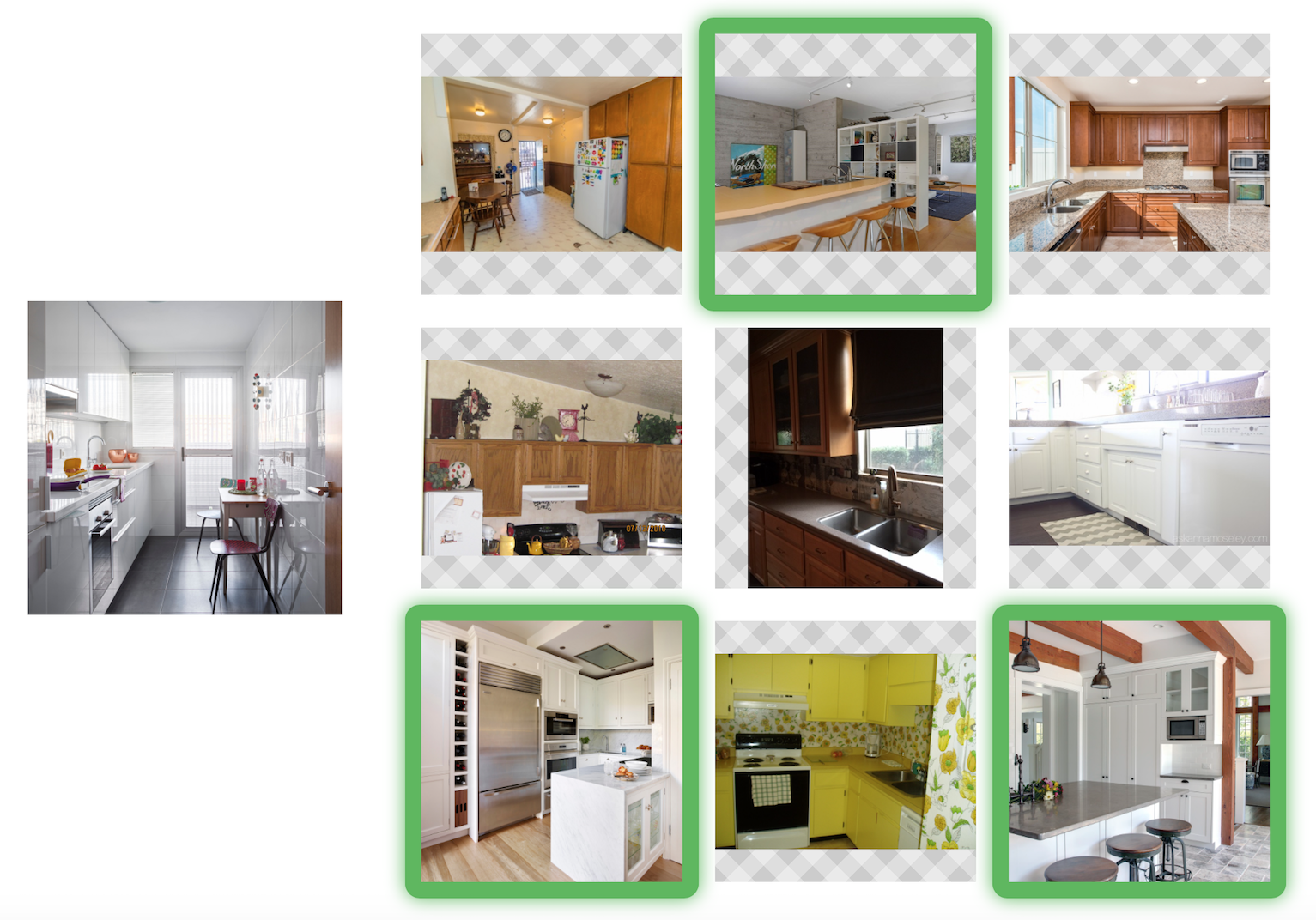}
\end{center}
   \caption{Crowdsourcing user interface for comparing photos based on their luxury level. Each probe image on the left is compared with 9 other images, uniformly drawn from the dataset. Using these comparisons, we obtain an embedding of real estate photos based on their luxury level and anchor images which represent different levels of luxury. \label{fig:GUI1}}
\end{figure*}
\begin{figure*}[!t]
\begin{center}
\begin{subfigure}{0.46\textwidth}
 \includegraphics[width=\linewidth]
 {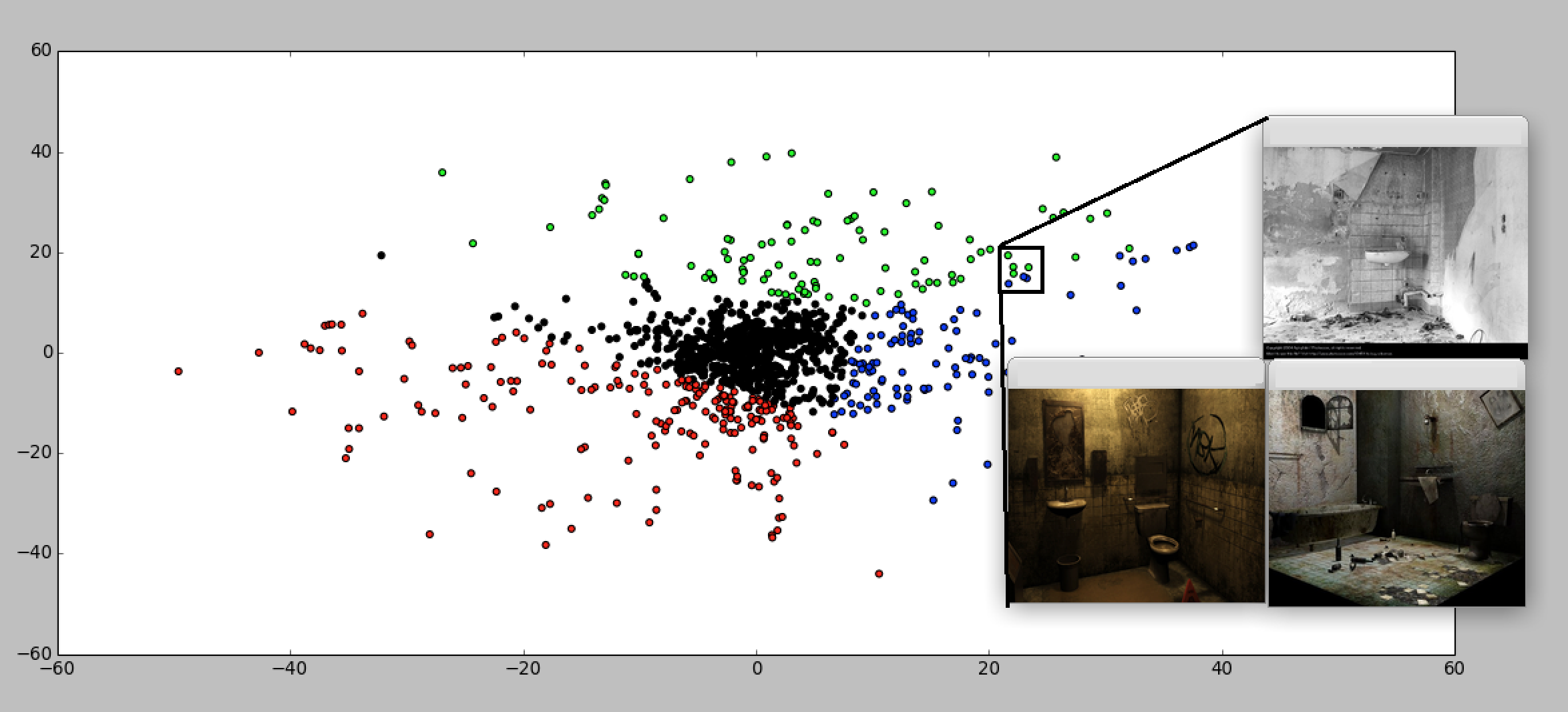}
\includegraphics[width=\linewidth]
{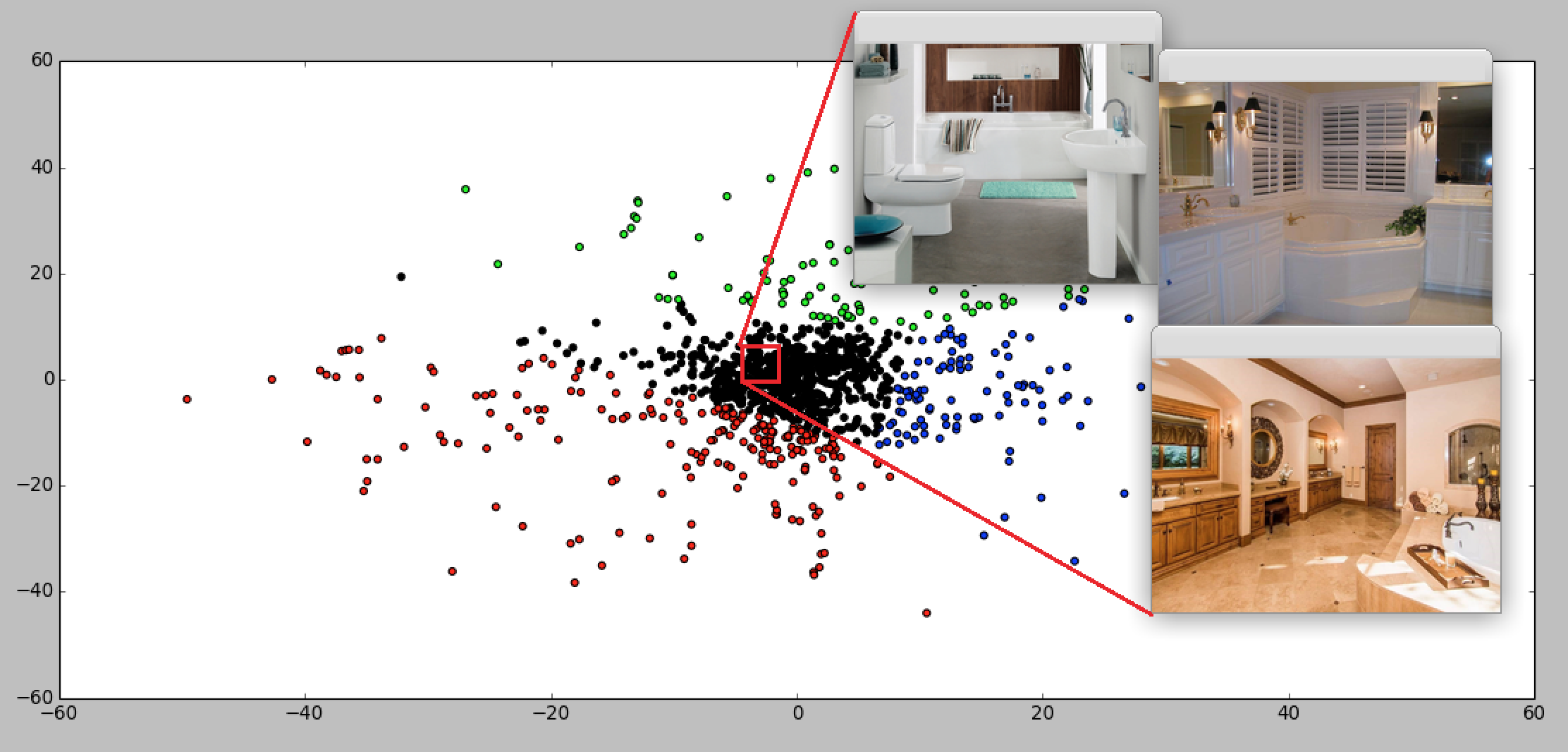}
  \caption{Bathroom}
\end{subfigure}
\hspace*{\fill}
\begin{subfigure}{0.46\textwidth}
\includegraphics[width=\linewidth]
{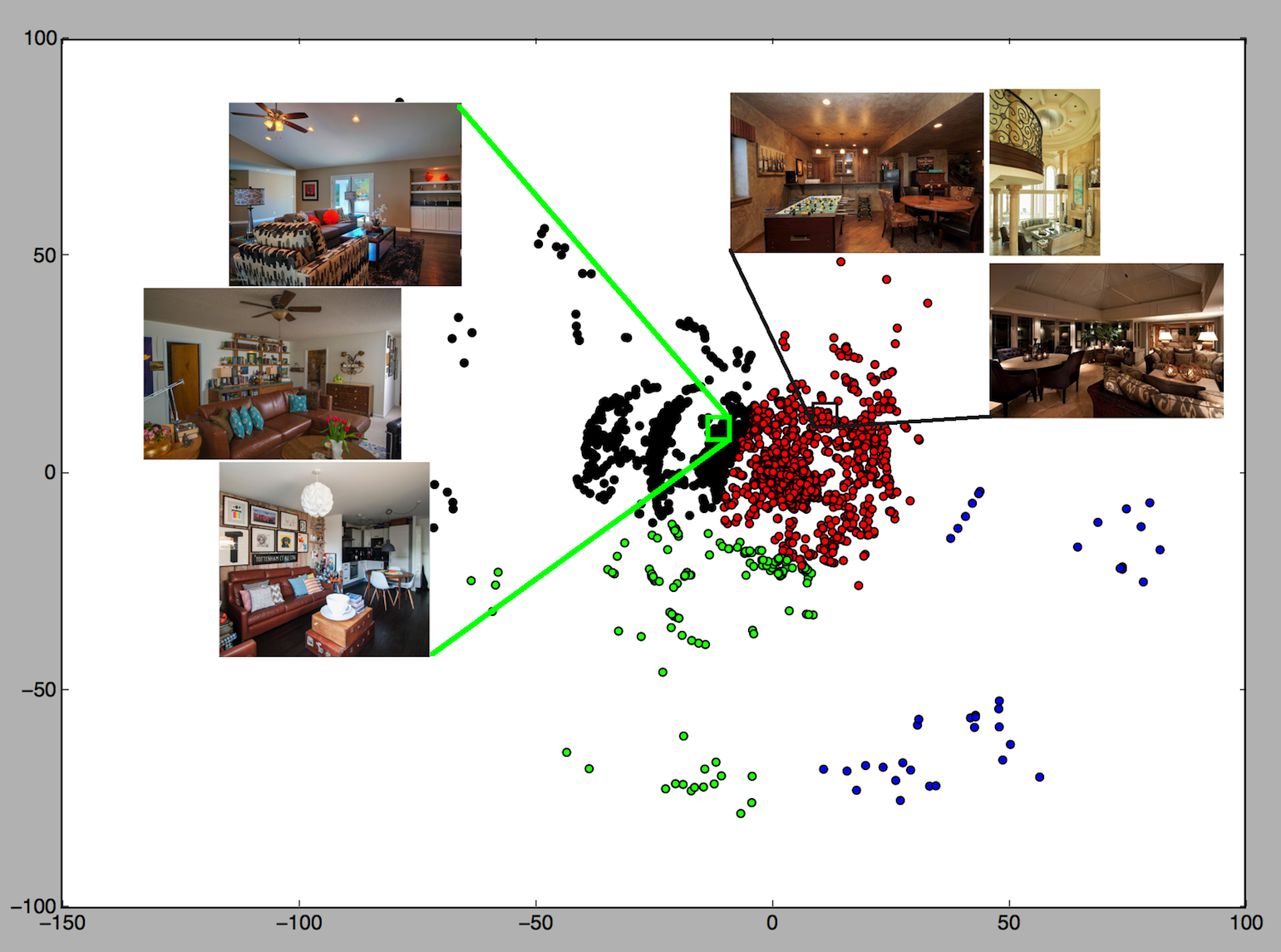}
 \caption{Living room}
\end{subfigure}

\begin{subfigure}{0.46\textwidth}
 \includegraphics[width=\linewidth]
 {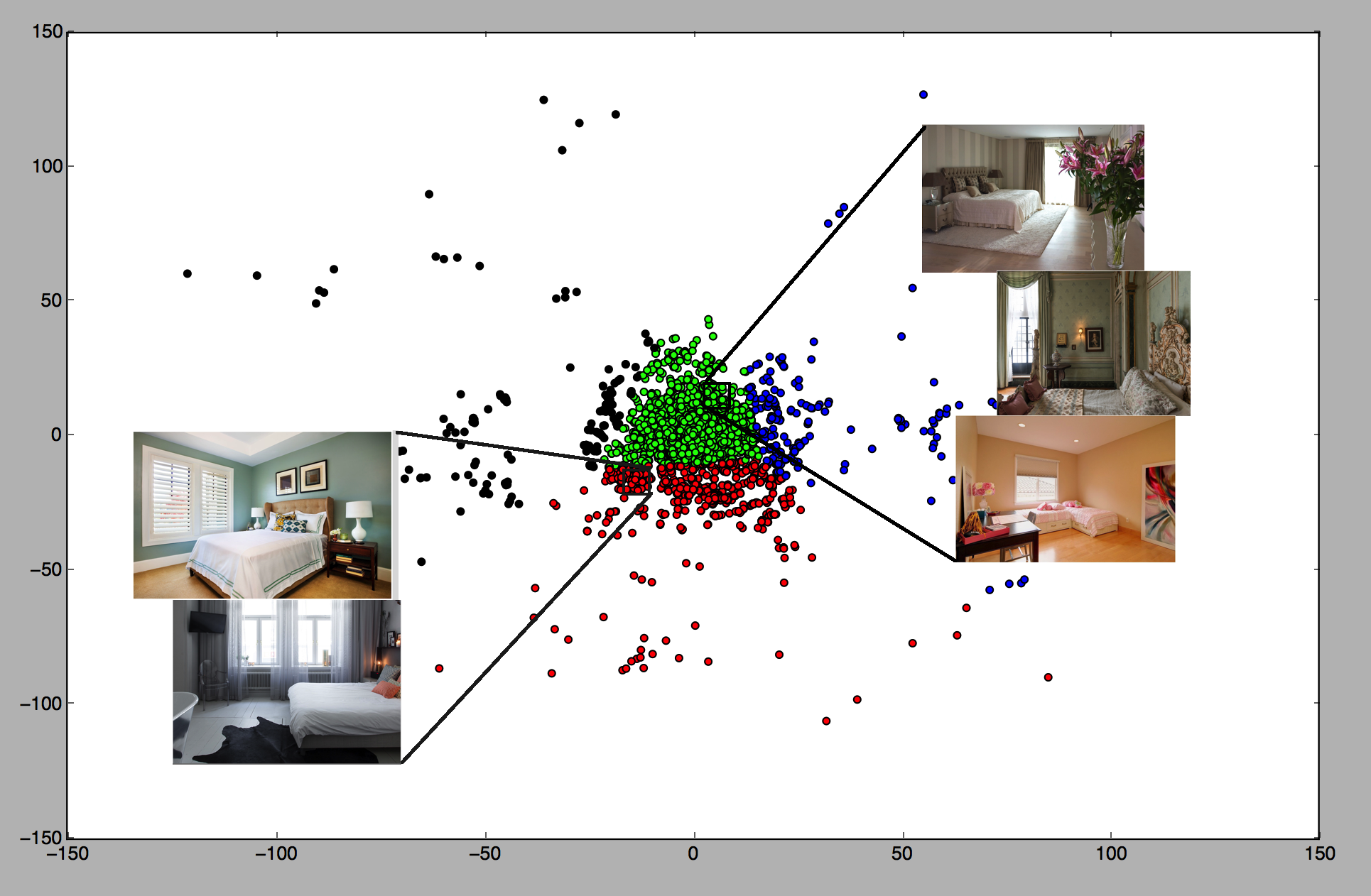}
  \caption{Bedroom}
\end{subfigure}
\hspace*{\fill} 
\begin{subfigure}{0.46\textwidth}
\includegraphics[width=\linewidth]      
{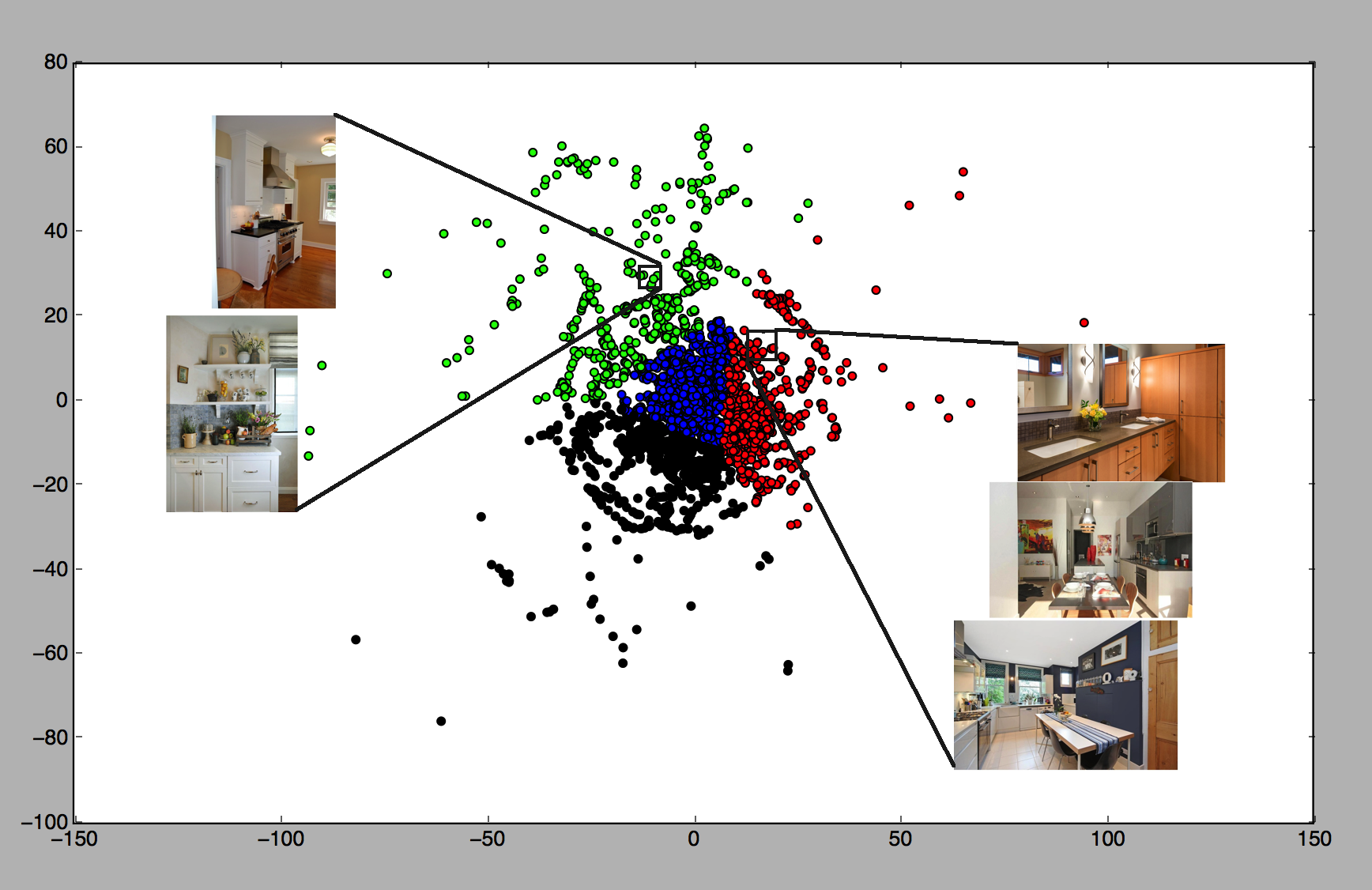}
  \caption{Kitchen}
\end{subfigure}
\end{center}
   \caption{2D embedding visualization of real estate photos based on their luxury using the t-STE algorithm. The embedding is obtained using 10,000 triplet comparisons. More luxurious photos are clustered at the center and more austere ones are scattered around.} 
\label{fig:embedding}
\end{figure*}
\subsection{Visual Design}

In spite of the importance of visual design and style, they are rarely addressed in the computer vision literature.
One of the main challenges in assigning a specific style to an image is that style is hard to define rigorously, as its interpretation can vary from person to person. In our work, we are interested in encoding information relevant to the luxury level of real estate photos. 

An approach for predicting the style of images is described in \cite{karayev2013recognizing}. It defines different types of image styles, and gathers a large-scale dataset of style-annotated photos that encompasses several different aspects of visual style. It also compares different image features for the task of style prediction and shows that features obtained from deep Convolutional Neural Networks (ConvNets) outperform other features. A visual search algorithm to match in-situ images with iconic product images is presented in \cite{bell2015learning}. It also provides an embedding that can be used for several visual search tasks including searching for products within a category, searching across categories, and searching for instances of a product in scenes. \cite{chechik2010large} presents a scalable algorithm for learning image similarity that captures both semantic and visual aspects of image similarity. \cite{ordonez2014furniture} discovers and categorizes learnable visual attributes from a large scale collection of images, tags and titles of furniture. A computational model of the recognition of real world scenes that bypasses
the segmentation and the processing of individual objects or regions is proposed in \cite{oliva2001modeling}. It is based on a low-dimensional
representation of the scene, called the Spatial Envelope. It proposes a set of perceptual dimensions that represent the dominant spatial structure of a scene.

\section{Our Approach}

In order to quantify the impact of visual characteristics on the value of residential properties, we need to encode real estate photos based on the value they add to the market price of a house. This value is tightly correlated with the concept of luxury. Luxurious designs increase the value of  a house, while austere ones decrease it. Hence, we focus on the problem of estimating the luxury level of real estate imagery and quantifying it in a way that can be used alongside the metadata to predict the price of residential properties. 

\subsection{Classifying Photos Based on Room Categories}

To make a reasonable comparison, we consider photos of each room type (kitchen, bathroom, etc.) separately. In other words, we expect that comparing rooms of the same type will give us better results than comparing different room categories. Hence, we trained a classifier to categorize pictures based on the categories shown in Table \ref{table:room}. 
In order to train the classifier, we used data from Places dataset \cite{zhou2014learning}, Houzz and Google image search. Our final dataset contains more than 200k images.

Using labeled pictures from our dataset, we trained DenseNet \cite{huang2016densely} for the task of classifying real estate photos to the following categories: bathroom, bedroom, kitchen, living room, dining room, interior (miscellaneous) and exterior.
Using this classifier, we achieved an accuracy of 91\% on the test set. After collecting a large dataset of real estate photos and metadata from Zillow, we applied the classifier to the images to categorize them based on their room type. Figure \ref{fig:room_classifier} shows examples of photos that are classified correctly and incorrectly. As we can observe from this figure, the classifier performs well for typical photos, while it sometimes wrongly classifies empty rooms and those rooms which combine elements from different categories.  

\subsection{Luxury Level Estimation}

After classifying the images based on their room categories, we used crowdsourcing for luxury level estimation. Since our goal is to quantify luxury level of photos, we need to assign a value to each photo to represent its level of luxury. Hence, we used a classification framework to categorize photos based on their luxury level. However, since it was not clear how many classes should be used and which photos should represent each of those classes, we used another crowdsourcing framework to compare images in our dataset according to their luxury level. Using these comparisons, we could obtain a low-dimensional embedding of real estate photos in which images with the same level of luxury are clustered near each other. 
By inspecting the embedding, we can determine the number of clusters that best represent variations in luxury, and choose the number of classes for the classification framework accordingly. We can also sample images from each cluster to represent different classes in the classification framework. 
\begin{figure*}[t]
\begin{center}
   \centering \includegraphics[width=0.88\linewidth]{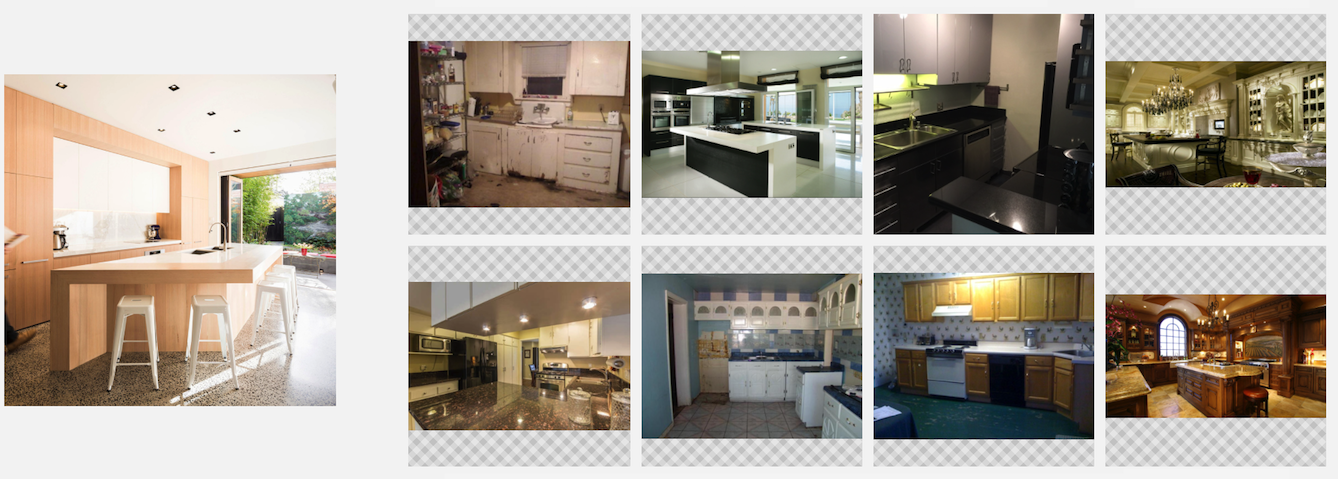}
\end{center}
   \caption{User interface for classifying real estate photos. Each of the 8 levels of luxury is represented with an anchor image, and the worker is asked to classify the probe image according to its luxury level.}
\label{fig:GUI2}
\end{figure*}
\begin{figure*}[!t]
\begin{center}
   \includegraphics[width=0.95\linewidth]{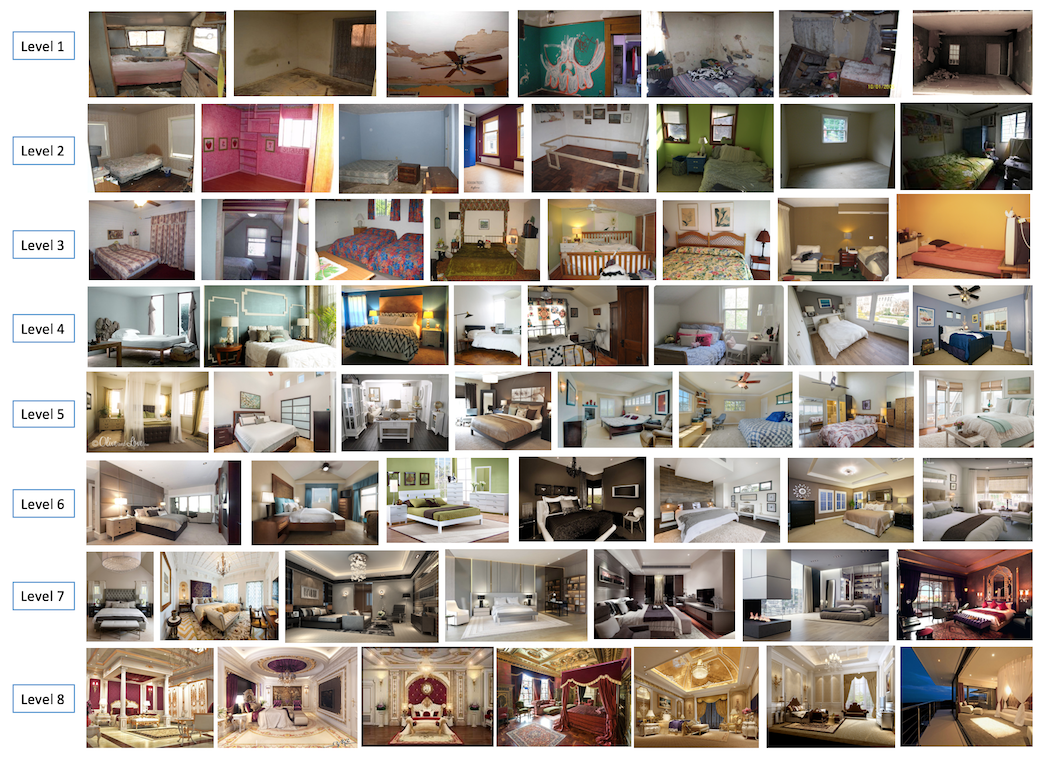}
\end{center}
   \caption{Examples of bedroom photos classified at different luxury levels. Level 1 represents the least level of luxury, and level 8 shows the highest.}
\label{fig:luxury_classifier}
\end{figure*}
\subsection{Crowdsourcing Framework}

We first discuss our crowdsourcing framework for comparing images based on their luxury. Motivated by \cite{wilber2014cost}, we presented a grid user interface to crowd workers, with a probe image on the left and 9 gallery images on the right. We asked the workers to select all images on the right that are similar in terms of luxury level to the image on the left. In order to extract meaningful comparisons from each grid, we want it to have images from several different luxury levels. Therefore, for each grid, we need to select images from our dataset uniformly.  

The images from Houzz have a `budget' label which determines the cost of each design. There are 4 different budget levels, and photos with a higher level represent more luxurious designs. Houses from Zillow are labeled with their offered price and Zestimates. We expect that houses with a higher price and Zestimate have more luxurious photos, and vice versa. Hence, to uniformly divide our dataset, we divided Zillow houses into 2 classes based on the average value of their offered prices and Zestimates. Moreover, to add images with low level of luxury to our dataset, we used Google image search. We searched for terms like `ugly bedroom,' `ugly kitchen,' etc. In this way, we obtained photos from ugly and spartan designs which generally decrease the value of a house.

In order to create each crowdsourcing grid, we sampled two images from each of the 2 classes of Zillow photos, four images from each of the 4 `budget' categories of Houzz pictures, two photos from the Places dataset, and two photos from Google search results. Then we selected one random picture as the probe and constructed the grid from the other 9 images. In this way, each grid contains photos of several different luxury levels to help the crowd workers provide meaningful comparisons among the pictures. A schematic of our crowdsourcing user interface is shown in figure~\ref{fig:GUI1}. We used Amazon Mechanical Turk (AMT) to collect comparisons on our images. 

\begin{figure*}[!t]
\begin{center}
   \includegraphics[width=\linewidth]{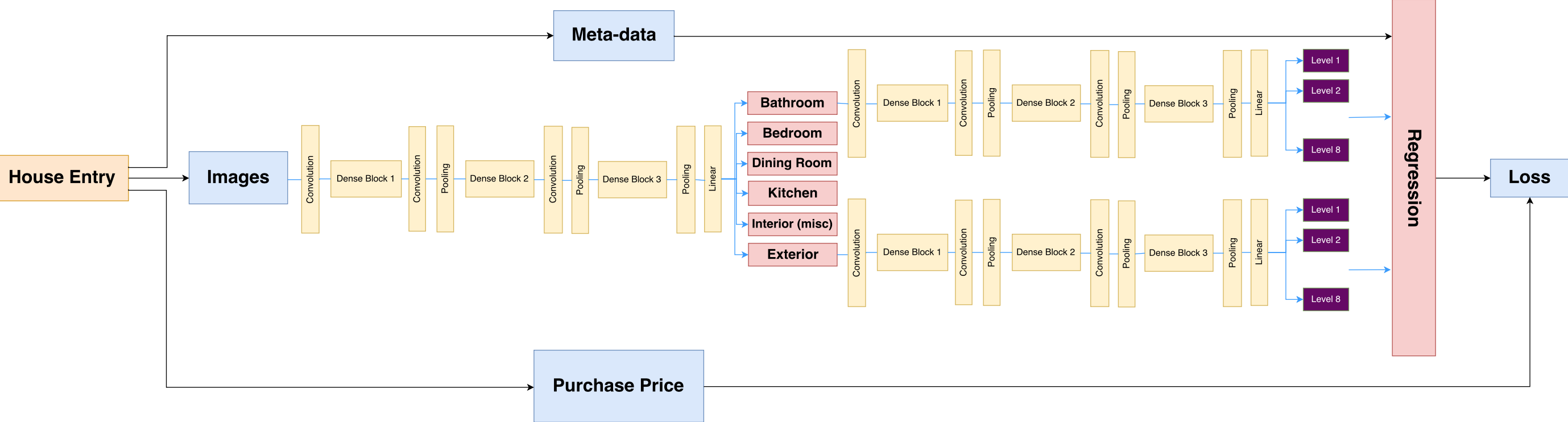}
\end{center}
   \caption{The price estimation network. After classifying photos based on their room category, 
 a vector representing luxury is extracted and concatenated with the normalized metadata vector and passed through a regression layer to produce the estimated price. The loss function is then computed as difference between the estimated price and the purchase price.}
\label{fig:price_estimation}
\end{figure*}

Using the crowdsourced data, we obtained triplet comparisons based on luxury level for a large-scale dataset of real estate photos. We then used the t-STE algorithm \cite{van2012stochastic} to obtain a two-dimensional embedding of the images. The result is shown in figure \ref{fig:embedding}. By examining the embedding, we observe that images with similar luxury levels are clustered near one another. This indicates the quality of the crowdsourced data.  
Each cluster represents a specific luxury level. 
We selected one anchor image from each cluster to represent photos in that cluster. 
Based on these representative pictures, we created another crowdsourcing task to rank photos according to their degree of luxury.  
Figure \ref{fig:GUI2} shows this task for kitchen images. Figure \ref{fig:luxury_classifier} illustrates examples of crowdsourcing results. It demonstrates that crowd workers generally performed well at categorizing photos based on their degree of luxury. This is due to the measures, such as tutorial rounds and catch trials, that we provided to ensure that workers comprehend the task and perform it attentively. 

\subsection{Price Estimation Network}

Using the classification framework shown in figure \ref{fig:GUI2}, we obtained luxury levels for a large training set of interior and exterior photos. We trained DenseNet \cite{huang2016densely} for the task of classifying real estate images based on their luxury level into 8 different categories. 
Then by using the trained classifier, we obtained levels of luxury for rooms of the houses in the Zillow dataset. For each house we obtained the average level of luxury for each of its room types. In this way, we obtained 7 values (one for each room type) representing luxury of each house. For houses with no photos of a specific room category, we used the average value of other categories to represent luxury level of that particular room. Then we concatenated the metadata vector with the vector representing the average luxury levels of rooms of the house. The metadata vector contains all the information about home characteristics like offered price, Zestimate, size, etc. Since different elements of the metadata vector (like offered price, age, number of bedrooms, etc.) are in different ranges, we first normalized the components. 
We computed the average value and the standard deviation of each component of the metadata vector for houses in our dataset. Then we used z-scoring for normalization: we subtracted the mean from each component and divided it by its standard deviation. 
In this way, the mean value of each entry of the metadata vector is approximately zero, and the standard deviation is approximately one. This allows us to obtain the function relating the market price of a house to its representative vector more easily from a computational point of view. 

The architecture for the price estimation network is shown in figure \ref{fig:price_estimation}. In order to train the network, we use the purchase price of recently sold houses as the ground-truth labels. As discussed in \cite{pagourtzi2003real}, the price on which the homeowner and the buyer agree best represents the market value of a house. As shown in the figure, for each house in the training dataset, we first classify its photos according to room type.
Pictures of each room type are then classified based on their luxury level. 
After extracting and normalizing the metadata vector of each house, we concatenate it with the vector that denotes the level of luxury. In this way, we obtain a representative vector for each house that captures the impact of its photos and metadata. 
Then we use kernel support vector regression (with Radial Basis Function as the kernel) to relate this vector to the actual value of the house. 
The input is the representative vector of each house and the output is the estimated price of the house. This price is then compared with the purchase price as the ground-truth. The difference between them represents the loss. Using the data from recently sold houses in our dataset, we obtained regression weights so as to minimize the loss function.

\section{Results and Discussion}
Using the crowdsourced data, we trained the price estimation network depicted in figure \ref{fig:price_estimation}. Then we used the network to estimate prices for a separate testing set of recently-sold houses. The error rate is determined by considering the difference between the price estimated by our network with the purchase price of houses in the test set. We evaluated the network's performance on a testing dataset of 1,000 recently-sold houses from Zillow. We compared the estimated price obtained using our network with the purchase price to find the error rate. The median error rate of our network is 5.8 percent which is better than the 7.9 percent median error rate of the Zestimate. Results are shown in Table \ref{table:Error}. Our automated valuation method improves upon that of Zillow by augmenting the input data with images. 

\subsection{Ablation Studies} We provide ablation analysis to evaluate effectiveness of different parts of the model. 
\begin{itemize}
\item We first consider using only the metadata and discarding visual information. The resulting median error rate is 8.0\%, which is very close to the Zestimate error rate. This is expected as Zestimate is included in the metadata. In other words, the regression network almost ignores all other elements in metadata except the Zestimate, which is optimized to be an accurate estimation of the price.   
\item In the next experiment, we remove the room classifier and train a single luxury level estimator for all rooms. As shown in Table \ref{table:Error}, the resulting median error rate is 6.7\%. The increase in error rate shows the importance of room classification as it helps the model focus on the fine-grained signal relevant to luxury within images of the same room type. 
\item We consider directly regressing the price. Instead of training the network for luxury estimation, we pre-train it on ImageNet. We use features extracted from the layer before the final classification layer to represent images. We train the regression network on these features and metadata. The resulting median error rate is 6.6\%. If we also fine-tune the feature extraction network, the error rate would be 6.4\%. The decreased performance is due to the fact that the amount of data containing the housing price is limited. This leads to features which do not correlate well with the value images add to the price of the house. However, we have a larger set of images without housing price values. As mentioned in the previous section, we annotate these images with luxury levels, and use that data to train a network for efficient luxury level estimation. This in turn helps for the price estimation since luxury is correlated with the additive value of images.    
\item After training each part of the model separately, we fine-tune the whole model end-to-end. This allows better information flow from the regression model to the luxury estimation network. As demonstrated in Table \ref{table:Error}, this leads to slight improvement (0.2\%) in the median error rate.    
\end{itemize}

\begin{table}[!t]
\caption{Median error rate of automated valuation methods}
\begin{center}
\begin{tabular}{c c}
Method & Median Error Rate \\
\specialrule{2.5pt}{1pt}{1pt}
Zestimate & $7.9\%$  \\
Ours (Vision-based) & $5.8\%$ \\
Only metadata & 8.0\% \\
Ours -- Room Classifier & {6.7\%} \\
Direct Regression & {6.6\%} \\
Direct Regression + Fine-tuning & {6.4\%} \\
Ours + Fine-tuning & {\bf 5.6\%} \\
\hline
\end{tabular}
\end{center}
\label{table:Error}
\end{table}

\section{Conclusion}
We have presented a novel algorithm to consider the impact of appearance on the value of residential properties. After collecting large datasets of real estate photos and metadata, we used a crowdsourcing pipeline to extract luxury-related information from interior and exterior photos. Based on the data we obtained via crowdsourcing, we trained a convolutional neural network to classify photos based on their level of luxury. 
Using purchase price of recently sold houses as their actual value, we trained a network to relate the market value of a house to its photos and metadata. We used our algorithm to estimate the price of houses in our dataset, and we showed that it provides a better value estimation than Zillow's estimates. Future avenues of research include assessing the effect of staging on the market value of a house, analyzing the impact of different furniture styles on the luxury level of real estate imagery, developing a user interface to help users select images that add more value to their residential properties, evaluating interior and exterior design photos from an aesthetic point of view, among others.

{\small
\bibliographystyle{ieee}
\bibliography{egbib}
}

\end{document}